\documentclass[11pt,a4paper]{article}
\usepackage[hyperref]{emnlp2020}
\usepackage{times}
\usepackage{latexsym}
\usepackage{booktabs}
\usepackage{graphicx}

\newcommand{\dimensions}{\textsc{about, as, to}} 
\newcommand{\newdataset}{\textsc{MDGender}} 

\def\Snospace~{\S{}} 

\usepackage{microtype}

\aclfinalcopy 

\title{Multi-Dimensional Gender Bias Classification} 

\author{
Emily Dinan$^*$, Angela Fan\thanks{~Joint first authors.} \dag, Ledell Wu, Jason Weston, Douwe Kiela, Adina Williams\\ 
Facebook AI Research \\ 
\dag Laboratoire Lorrain d'Informatique et 
Applications (LORIA)
}

\begin{document}
\maketitle
\begin{abstract}
Machine learning models are trained to find patterns in data. NLP models can inadvertently learn socially undesirable patterns when training on gender biased text. In this work, we propose a general framework that decomposes gender bias in text along several pragmatic and semantic dimensions: bias from the gender of the person being spoken about, bias from the gender of the person being spoken to, and bias from the gender of the speaker. Using this fine-grained framework, we automatically annotate eight large scale datasets with gender information. In addition, we collect a novel, crowdsourced evaluation benchmark of utterance-level gender rewrites.  Distinguishing between gender bias along multiple dimensions is important,
as it enables us to train finer-grained gender bias classifiers. We show our classifiers prove valuable for a variety of important applications, such as controlling for gender bias in generative
models, detecting gender bias in arbitrary text, and shed light on offensive language in terms of genderedness.
\end{abstract}

\section{Introduction} \label{sec:intro}

Language is a social behavior, and as such, it is a primary means by which people communicate, express their identities, and socially categorize themselves and others. Such social information is present in the words we write and, consequently, in the text we use to train our NLP models. In particular, models often can unwittingly learn negative associations about protected groups present in their training data and propagate them. In particular, NLP models often learn biases against others based on their gender \citep{bolukbasi2016man,hovy2016, caliskan2017, rudinger-etal-2017-social,garg2018, gonen2019, dinan2019queens}. Since unwanted gender biases can affect downstream applications---sometimes even leading to poor user experiences---understanding and mitigating gender bias is an important step towards making NLP tools and models safer, more equitable, and more fair. We provide a finer-grained framework for this purpose, analyze the presence of gender bias in models and data, and empower others by releasing tools that can be employed to address these issues for numerous text-based use-cases.

\begin{figure}[t]
    \centering
    \includegraphics[width=\columnwidth]{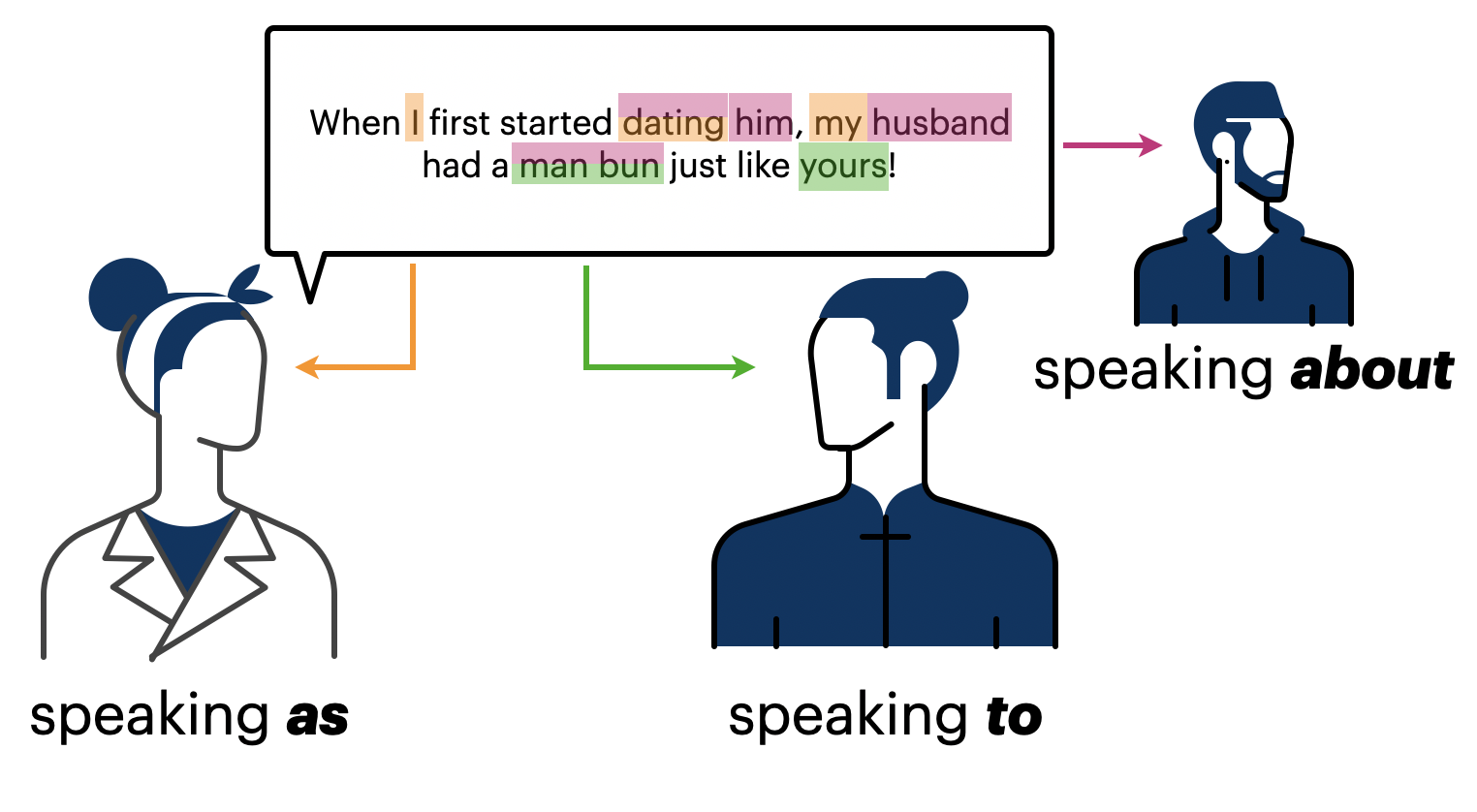}
    \caption{\textbf{Framework for Gender Bias in Dialogue}. We propose a framework separating gendered language based on who you are speaking \textsc{about}, speaking \textsc{to}, and speaking \textsc{as}.}
    \label{fig:frameworkfig}
\end{figure}

While many works have explored methods for removing gender bias from text \citep{bolukbasi2016man,emami2019,counterfactualDataSubstitution, dinan2019queens, kaneko2019gender,zmigrod2019, ravfogel2020}, no extant work on classifying gender or removing gender bias has incorporated facts about how humans collaboratively and socially construct our language and identities. We propose a pragmatic and semantic framework for measuring bias along three dimensions that builds on knowledge of the conversational and performative aspects of gender, as illustrated in \autoref{fig:frameworkfig}. Recognizing these dimensions is important, because gender along each dimension can affect text differently, for example, by modifying word choice or imposing different preferences in how we construct sentences. 

Decomposing gender into separate dimensions also allows for better identification of gender bias, which subsequently enables us to train a suite of classifiers for detecting different kinds of gender bias in text. We train several classifiers on freely available data that we annotate with gender information along our dimensions. We also collect a new crowdsourced dataset (\newdataset) for better evaluation of gender classifier performance. The classifiers we train have a wide variety of potential applications.  We evaluate them on three: controlling the genderedness of generated text, detecting gendered text, and examining the relationship between gender bias and offensive language. In addition, we expect them to be useful in future for many text applications such as detecting gender imbalance in newly created training corpora or model-generated text.

In this work, we make four main contributions: we propose a multi-dimensional framework (\dimensions) for measuring and mitigating gender bias in language and NLP models, we introduce an evaluation dataset for performing gender identification that contains utterances re-written from the perspective of a specific gender along all three dimensions, we train a suite of classifiers capable of labeling gender in both a single and multitask set up, and finally we illustrate our classifiers' utility for several downstream applications. All datasets, annotations, and classifiers will be released publicly to facilitate further research into the important problem of gender bias in language. 

\section{Related Work}\label{sec:relwork}

Gender affects myriad aspects of NLP, including corpora, tasks, algorithms, and systems \citep{chang2019bias,costa2019,sun-etal-2019-mitigating}. For example, statistical gender biases are rampant in word embeddings \citep{jurgens-etal-2012-semeval, bolukbasi2016man, caliskan2017, garg2018, zhao-etal-2018-learning, basta2019, chaloner2019, du2019, gonen2019,kaneko2019gender,zhao-etal-2019-gender}---even multilingual ones \citep{gonen2019does, zhou2019}---and affect a wide range of downstream tasks including coreference resolution \citep{zhao-etal-2018-gender,cao2019toward, emami2019}, part-of-speech and dependency parsing \citep{garimella2019}, unigram language modeling \citep{qian2019reducing}, appropriate turn-taking classification \citep{lepp2019pardon}, relation extraction \citep{gaut2019towards}, identification of offensive content \citep{sharifirad2019using,sharifirad2019learning}, and machine translation \citep{stanovsky2019evaluating}. Furthermore, translations are judged as having been produced by older and more male speakers than the original was \citep{hovy2020}.  

For dialogue text particularly, gender biases in training corpora have been found to be amplified in machine learning models \citep{lee2019exploring, dinan2019queens, dialoguefairness}. 
While many of the works cited above propose methods of mitigating the unwanted effects of gender on text, \newcite{counterfactualDataSubstitution,zmigrod2019, dinan2019queens} in particular rely on counterfactual data to alter the training distribution to offset gender-based statistical imbalances (see \autoref{subsec:models} for more discussion of training set imbalances). Also relevant is \newcite[PASTEL]{kang2019}, which introduces a parallel style corpus and shows gains on style-transfer across binary genders. In this work, we provide a clean new way to understand gender bias that extends to the dialogue use-case by independently investigating the contribution of author gender to data created by humans. 

Most relevant to this work, \newcite{sap2019social} proposes a framework for modeling pragmatic aspects of many social biases in text, such as  intent to offend, for guiding discovery of new instances of social bias. These works focus on complementary aspects of a larger goal---namely, making NLP safe and inclusive for everyone---but they differ in several ways. Here, we treat statistical gender bias in human or model generated text  specifically, allotting it the focused and nuanced attention that such a complicated phenomenon deserves. \newcite{sap2019social} takes a different perspective, and aims to characterize the broader landscape of negative stereotypes in social media text, an approach which can make parallels apparent across different types of socially harmful content. Moreover, they consider different pragmatic dimensions than we do: they target negatively stereotyped  commonsense implications in arguably innocuous statements, whereas we investigate pragmatic dimensions that straightforwardly map to conversational roles (i.e., topics, addressees, and authors of content). As such, we believe the two frameworks to be fully compatible.

Also relevant is the \textbf{intersectionality} of gender identity, i.e., when gender non-additively interacts with other identity characteristics. Negative gender stereotyping is known to be weakened or reinforced by the presence of other social factors, such as dialect \citep{tatman-2017-gender}, class \citep{degaetano-ortlieb-2018-stylistic} and race \citep{crenshaw1989}. These differences have been found to affect gender classification in images \citep{pmlr-v81-buolamwini18a}, and also in sentences encoders \citep{may-etal-2019-measuring}. We acknowledge that these are crucial considerations, but set them aside for follow-up work.

\section{Dimensions of Gender Bias}\label{sec:background}

Gender infiltrates language differently depending on the conversational role played by the people using that language (see \autoref{fig:frameworkfig}). We propose a framework for decomposing gender bias into three separate dimensions: bias when speaking \textsc{about} someone, bias when speaking \textsc{to} someone, and bias from speaking \textsc{as} someone. In this section, we first define \textit{bias} and \textit{gender}, and then motivate and describe our three dimensions. 

\begin{table*}
\setlength\tabcolsep{1.5pt}
\begin{center}
\small
\begin{tabular}{ccccccccc}
\toprule
\multicolumn{3}{c}{\sc Verbs} &  \multicolumn{3}{c}{\sc Adjectives} & \multicolumn{3}{c}{\sc Adverbs}\\
\cmidrule(lr){1-3}\cmidrule(lr){4-6}\cmidrule(lr){7-9}
M & F & N & M & F & N & M & F & N \\
\cmidrule(lr){1-1}\cmidrule(lr){2-2}\cmidrule(lr){3-3}
\cmidrule(lr){4-4}\cmidrule(lr){5-5}\cmidrule(lr){6-6}
\cmidrule(lr){7-7}\cmidrule(lr){8-8}\cmidrule(lr){9-9}
finance & steamed & increases & akin & feminist & optional & ethnically & romantically & westward \\
presiding & actor & range & vain & lesbian & tropical & intimately & aground & inland \\
oversee & kisses & dissipated & descriptive & uneven & volcanic & soundly & emotionally & low \\
survives & towed & vary & bench & transgender & glacial & upstairs & sexually & automatically \\
disagreed & guest & engined & sicilian & feminine & abundant & alongside & happily & typically \\
obliged & modelling & tailed & 24-hour & female & variable & artistically & socially & faster \\
filling & cooking & excavated & optimistic & reproductive & malay & randomly & anymore & normally \\
reassigned & kissing & forested & weird & sexy & overhead & hotly & really & round \\
pledged & danced & upgraded & ordained & blonde & variant & lesser & positively & usually \\
agreeing & studies & electrified & factual & pregnant & sandy & convincingly & incredibly & slightly \\
\bottomrule
\end{tabular}
\end{center}
\caption{\textbf{Bias in Wikipedia}. We look at the most over-represented words in biographies of men and women, respectively, in Wikipedia. We also compare to a set of over-represented words in gender-neutral pages. We use a part-of-speech tagger \cite{spacy2} and limit our analysis to words that appear at least $500$ times.}
\label{table:wikipedia-bias}
\end{table*}

\subsection{Definitions of Bias and Gender}
\paragraph{Bias.} 

In an ideal world, we would expect little difference between texts describing men, women, and people with other gender identities, aside from the use of explicitly gendered words, like pronouns or names. A machine learning model, then, would be unable to pick up on statistical differences among gender labels (i.e., gender \textbf{bias}), because such differences would not exist.  Unfortunately, we know this is not the case. For example, Table~\ref{table:wikipedia-bias} provides examples of adjectives, adverbs, and verbs that are more common in Wikipedia biographies of people of certain genders. This list was generated by counting all verbs, adjectives, and adverbs (using a part-of-speech tagger from \citet{spacy2}) that appear in a large section of biographies of Wikipedia.  We then computed $P(\mbox{word}~|~\mbox{gender})/P(\mbox{word})$ for words that appear more than $500$ times. The top over-represented verbs, adjectives, and adverbs using this calculated metric are displayed for each gender.

In an imagined future, a classifier trained to identify gendered text would have (close to) random performance on non-gender-biased future data, because the future would be free of the statistical biases plaguing current-day data. These statistical biases are what make it possible for current-day classifiers to perform better than random chance. We know that current-day classifiers are gender biased, because they achieve much better than random performance by learning distributional differences in how current-day texts use gender; we show this in  \autoref{sec:results}.  
These classifiers learn to pick up on these statistical biases in text \emph{in addition to} explicit gender markers (like \textit{she}).\footnote{We caution the reader that ``the term bias is often used to refer to demographic disparities in algorithmic systems that are objectionable for societal reasons'' \citep[14]{FAIRML}; we restrict our use of \textbf{bias} to its traditional definition here.}

\paragraph{Gender.} Gender manifests itself in language in numerous ways. In this work, we are interested in gender as it is used in English when referring to people and other sentient agents, or when discussing their identities, actions, or behaviors. We annotate gender with four potential values: \textit{masculine, feminine, neutral} and \textit{unknown} --- which allows us to go beyond the oppositional male-female gender binary. We take the \textit{neutral} category to contain characters with either non-binary gender identity, or an identity which is unspecified for gender by definition (say, for a magic tree).\footnote{We fully acknowledge the existence and importance of all chosen gender identities---including, but not limited to non-binary, gender fluid, poly-gender, pan-gender, alia-gender, agender---for the end goal of achieving accessible, inclusive, and fair NLP. However, these topics require a more nuanced investigation than is feasible using na\"ive crowdworkers.} We also include an \textit{unknown} category for when there might be a gender identity at play, but the gender is not known or readily inferrable by crowdworkers from the text (e.g., in English, one would not be able to infer gender from just the short text ``Hello!'').

\subsection{Gender in Multiple Dimensions}\label{subsec:dimensions}






Exploring gender's influence on language has been a fruitful and active area of research in many disciplines, each of which brings its own unique perspectives to the topic \citep{lakoff1973,butler1990, cameron1990, lakoff1990,swann1992,crawford1995,weatherall2002, sunderland2006, eckert2013, mills2014, coates2015, talbot2019}. In this section, we propose a framework  that decomposes gender's contribution along three conversational dimensions to enable finer-grained classification of gender's effects on text from multiple domains. 

\paragraph{\textit{Speaking About}:\ Gender of the Topic.}
It's well known that we change how we speak about others depending on who they are \citep{hymes1974,rickford1994}, and, in particular, based on their gender \citep{lakoff1973}. People often change how they refer to others depending on the gender identity of the individual being spoken about \citep{eckert1992}. For example, adjectives which describe women have been shown to differ from those used to describe men in numerous situations \citep{trix2003, gaucher2011, moon2014, hoyle2019}, as do verbs that take nouns referring to men as opposed to women \cite{guerin1994, hoyle2019}. Furthermore, metaphorical extensions---which can shed light on how we construct conceptual categories \citep{lakoff1980}---to men and women starkly differ (\citealt{fontecha2003}; \citealt[325]{holmes2013}; \citealt{amery2015}). 

\paragraph{\textit{Speaking To}:\ Gender of the Addressee.}
People often adjust their speech based on who they are speaking with---their addressee(s)---to show solidarity with their audience or express social distance \cite{wish1976,bell1984language, hovy1987, rickford1994, bell1997, eckert2001}. We expect the addressee's gender to affect, for example, the way a man might communicate with another man about styling their hair would differ from how he might communicate with a woman about the same topic. 

\paragraph{\textit{Speaking As}:\ Gender of the Speaker.}
People react to content differently depending on who created it.\footnote{We will interchangeably use the terms \textbf{speaker} and \textbf{author} here to refer to a creator of textual content throughout.} For example, \newcite{sap2019risk} find that na\"\i ve annotators are much less likely to flag as offensive certain content referring to race, if they have been told the author of that content speaks a dialect that signals in-group membership (i.e., is less likely to be intended to offend). Like race, gender is often described as a ``fundamental'' category for self-identification and self-description \citep[315]{banaji1994}, with men, women, and non-binary people differing in how they actively create and perceive of their own gender identities \citep{west1987}. Who someone is \textit{speaking as} strongly affect what they may say and how they say it, down to the level of their choices of adjectives and verbs in self-descriptions \citep{charyton2007, wetzel2012}. Even children as young as two dislike when adults misattribute a gender to them \citep{money1972,bussey1986}, suggesting that gender is indeed an important component of identity.

Our \textit{Speaking As} dimension builds on prior work on author attribution, a concept purported to hail from English logician Augustus de Morgan \citep{mendenhall1887characteristic}, who suggested that authors could be distinguished based on the average word length of their texts. Since then, sample statistics and NLP tools have been used for applications such as settling authorship disputes \citep{mosteller1984}, forensic investigations \citep{frantzeskou2006, rocha2016, peng2016}, or extracting a stylistic fingerprint from text that enables the author to be identified \citep{stamatatosetal1999automatic, luyckx2008, argamon2009, stamatatos2009, raghavanetal2010authorship, cheng2011, stamatatos-2017-authorship}. 
More specifically, automatic gender attribution has reported many successes \citep{koppel2002, koolenvancranenburgh2017stereotypes, qian2019gender}, often driven by the fact that authors of specific genders tend to prefer producing content about topics that belie those gender \citep{sarawgietal2011gender}. Given this, we might additionally expect differences between genders along our \textit{Speaking As} and \textit{Speaking About} dimensions to interact, further motivating them as separate dimensions.

\begin{table}
\centering
\small 
\begin{tabular}{lllllll}
\toprule
\textbf{Dataset} & M & F & N & U  & \textbf{Dim} \\
\midrule
\multicolumn{5}{l}{\emph{Training Data}} \\
\midrule
Wikipedia & 10M & 1M & 1M & - & \textsc{about} \\
Image Chat & 39K & 15K & 154K & - & \textsc{about} \\
Funpedia & 19K & 3K & 1K & - & \textsc{about} \\
Wizard & 6K & 1K & 1K & - & \textsc{about} \\
Yelp & 1M & 1M & - & - & \textsc{as}\\
ConvAI2 & 22K & 22K & - & 86K & \textsc{as}\\
ConvAI2 & 22K & 22K & - & 86K & \textsc{to}\\
OpenSub & 149K & 69K & - & 131K & \textsc{as}\\
OpenSub & 95K & 45K & - & 209K & \textsc{to}\\
LIGHT & 13K & 8K & -& 83K & \textsc{as}\\
LIGHT & 13K & 8K & - & 83K & \textsc{to}\\
\midrule 
\multicolumn{5}{l}{\emph{Evaluation Data}} \\
\midrule 
\newdataset & 384 & 401 & - & - & \textsc{about}\\
\newdataset & 396 & 371 & -  & - & \textsc{as}\\
\newdataset & 411 & 382 & - & - & \textsc{to}\\
\bottomrule
\end{tabular}
\caption{\label{table:data_stats} \textbf{Dataset Statistics}. Dataset statistics on the eight training datasets and new evaluation dataset, \newdataset with respect to each label.}
\end{table}

\section{Creating Gender Classifiers}

\begin{table*}
\centering
\setlength\tabcolsep{5pt}
\begin{tabular}{lllllllllllll}
\toprule
Model & \multicolumn{3}{c}{\textbf{About}} & \multicolumn{3}{c}{\textbf{To}} & \multicolumn{3}{c}{\textbf{As}}  \\ 
\cmidrule(lr){2-4}\cmidrule(lr){5-7}\cmidrule(lr){8-10}
      & Avg. & M & F                      &   Avg. & M  & F                  &  Avg. & M & F & All Avg. \\
\midrule 
SingleTask \textsc{about} & \textbf{70.43} & 63.54 & 77.31 & 44.44 & 36.25 & 52.62 & 67.75 & 69.19 & 66.31  & 60.87 \\
SingleTask \textsc{to}  & 50.12 & 99.74 & 0.5 & 49.39 & 95.38 & 3.4 & 50.41 & 100 & 0.81  & 49.97 \\
SingleTask \textsc{as} & 46.97 & 51.3 & 42.4 & 57.27 & 67.15 & 47.38 & \textbf{78.21} & 70.71 & 85.71  & 60.82 \\ 
MultiTask  & 62.59 & 64.32 & 60.85 & \textbf{78.25} &  73.24 & 83.25  & 72.15 & 66.67 & 77.63 & \textbf{67.13}  \\ 
\bottomrule
\end{tabular}
\caption{\label{table-one} \textbf{Accuracy on the novel evaluation dataset \newdataset} comparing single task classifiers to our multi-task classifiers. We report accuracy on the \emph{masculine} and the \emph{feminine} classes, as well as the average of these two metrics. Finally, we report the average (of the M-F averages) across the three dimensions. \newdataset was collected to enable evaluation on the \emph{masculine} and \emph{femninine} classes, for which much of the training data is noisy.}
\end{table*}

\begin{table}
\centering
\scalebox{0.95}{
\setlength\tabcolsep{3pt}
\begin{tabular}{lccccc}
\toprule
Model & \multicolumn{5}{c}{\textbf{Multitask Performance}} \\ 
      &  M & F & N & Avg. & Dim.\\
\midrule 
Wikipedia & 87.4 & 86.65 & 55.2 & 77.22 & \sc{about} \\ 
Image Chat & 36.48 & 83.56 & 33.22 & 51.09 & \sc{about} \\
Funpedia & 75.82 & 82.24 & 70.52 & 76.2 & \sc{about} \\ 
Wizard & 64.51 & 83.33 & 81.82 &76.55 & \sc{about} \\ 
Yelp & 73.92 & 65.08 & - & 69.5 & \sc{as} \\ 
ConvAI2 & 44 & 65.65 & - & 54.83 & \sc{as} \\ 
ConvAI2 &  45.98 & 61.28 & - & 53.63 & \sc{to} \\ 
OpenSubtitles & 56.95 & 59.31 & - & 58.12 & \sc{as} \\ 
OpenSubtitles & 53.73 & 60.29 & - & 57.01 & \sc{to} \\ 
LIGHT & 51.57 & 65.72 & - & 58.65 & \sc{as} \\ 
LIGHT & 51.92 & 68.48 & - & 60.2 & \sc{to} \\
\bottomrule
\end{tabular}
}
\caption{\label{table-two} \textbf{Performance of the multitask model on the test sets from our training data.}~We evaluate the multi-task model on the test sets for the training datasets. We report accuracy on each (gold) label---masculine, feminine, and neutral---and the average of the three. We do not report accuracy on imputed labels.}
\end{table}

Previous work on gender bias classification has been predominantly single-task---often supervised on the task of analogy---and relied mainly on word lists, that are binarily \cite{bolukbasi2016man,zhao-etal-2018-learning,zhao-etal-2019-gender, gonen2019}---and sometimes also explicitly \citep{caliskan2017,hoyle2019}---gendered. While wordlist-based approaches provided a solid start on attacking the problem of gender bias, they are insufficient for multiple reasons. First, they conflate different conversational dimensions of gender bias, and are therefore unable to detect the subtle pragmatic differences that are of interest here. Further, all existing gendered word lists for English are limited, by construction, to explicitly binarily gendered words (e.g., \textit{sister} vs. \textit{brother}). Not only is binary gender wholly inadequate for the task, but restricting to explicitly gendered words is itself problematic, since we know that many words aren't explicitly gendered, but \emph{are} strongly statistically gendered (see \autoref{table:wikipedia-bias}). Rather than solely relying on a brittle binary gender label from a global word list, our approach will also allow for gender bias to be determined flexibly over multiple words in context (Note: this will be crucial for examples that only receive gendered interpretations when in a particular context; for example, `bag' disparagingly refers to an elderly woman, but only in the context of `old', and `cup' hints at masculine gender only in the context of `wear'). 

Instead, we develop classifiers that can decompose gender bias over full sentences into semantic and/or pragmatic dimensions (\emph{about/to/as}), additionally including gender information that (i) falls outside the male-female binary, (ii) can be contextually determined, and (iii) is statistically as opposed to explicitly gendered. In the subsequent sections, we provide details for training these classifiers as well as details regarding the annotation of data for such training.

\subsection{Models}\label{subsec:models}
We outline how these classifiers are trained  to predict gender bias along the three dimensions, providing details of the classifier architectures as well as how the data labels are used. We train single-task and a multi-task classifiers for different purposes: the former will leverage gender information from each contextual dimension individually, and the latter should have broad applicability across all three.

\paragraph{Single Task Setting.} In the single-task setting, we predict \textit{masculine}, \textit{feminine}, or  \textit{neutral} for each dimension -- allowing the classifier to predict any of the three labels for the \textit{unknown} category). 

\paragraph{Multitask Setting.} To obtain a classifier capable of multi-tasking across the \emph{about/to/as} dimensions, we train a model to score and rank a set of possible classes given textual input. For example, if given \emph{Hey, John, I'm Jane!}, the model is trained to rank elements of both the sets \{TO:masculine, TO:feminine,  TO:neutral\} and \{AS:masculine, AS:feminine, AS:neutral\} and produce appropriate labels TO:masculine and AS:feminine. Models are trained and evaluated on the annotated datasets. 

\paragraph{Model Architectures.} For single task and multitask models, we use a pretrained Transformer \citep{vaswani2017attention} to find representations for the textual input and set of classes. Classes are scored---and then ranked---by taking a dot product between the representations of the textual input and a given class, following the bi-encoder architecture \citep{humeau2019real} trained with cross entropy. The same architecture and pre-training as BERT \citep{Devlin2018BERTPO} are used throughout.  We use \texttt{ParlAI} for model training \citep{miller2017parlai}. We will release data and models. 

\paragraph{Model Labels.} Many of our annotated datasets contain cases where the \textsc{about, as, to} labels are unknown. We retain these examples during training, but use two techniques to handle them. 
If the true label is \emph{unknown} (for example, in Wikipedia, we do not know the gender of the author, so the \emph{as} dimension is unknown), we either impute it or provide a label at random. For data for which the \emph{about} label is unknown, we impute it using a classifier trained only on data for which this label is present. For data for which the \emph{to} or \emph{as} label is unknown, we provide a label at random, choosing between \emph{masculine} and \emph{feminine}. From epoch to epoch, we switch these arbitrarily assigned labels so that the model learns to assign the \emph{masculine} and \emph{feminine} labels with roughly equal probability to examples for which the gender is \emph{unknown}. This label flipping allows us to retain greater quantities of data by preserving unknown samples. Additionally, we note that during training, we balance the data across the \emph{masculine}, \emph{feminine}, and \emph{neutral} classes by oversampling from classes with fewer examples. We do this because much of the data is highly imbalanced: for example, over $>80\%$ of examples from Wikipedia are labeled \emph{masculine} (\autoref{table:data_stats}). We also early stop on the average accuracy across all three classes.

\subsection{Data}
Next, we describe how we annotated our training data, including both the 8 existing datasets and our novel evaluation dataset, \newdataset.

\paragraph{Annotation of Existing Datasets.}

To enable training our classifiers, we leverage a variety of existing datasets. Since one of our main contributions is a suite of open-source general-purpose gender bias classifiers, we selected datasets for training based on three criteria: inclusion of recoverable information about one or more of our dimensions, diversity in textual domain, and high quality, open data use. Once we narrowed our search to free, open source and freely available datasets, we maximized domain diversity by selecting datasets with high quality annotations along at least one of our dimensions (e.g., dialogue datasets have information on author and addressee gender, biographies have information on topic gender, and restaurant reviews have information on author gender). 

The datasets are: Wikipedia, Funpedia (a less formal version of Wikipedia) \cite{miller2017parlai}, Wizard of Wikipedia (knowledge-based conversation) \cite{dinan2019wizard}, Yelp Reviews\footnote{\url{https://yelp.com/dataset}}, ConvAI2 (chit-chat dialogue) \cite{dinan2019second}, ImageChat (chit-chat dialogue about an image) \cite{shuster2018imagechat}, OpenSubtitles (dialogue from movies) \cite{lison2016opensubtitles2016}, and LIGHT (chit-chat fantasy dialogue) \cite{urbanek2019light}. We use data from multiple domains to represent different styles of text---from formal writing to chitchat---and different vocabularies. Further, several datasets are known to contain statistical imbalances and biases with regards to how people of different genders are described and represented, such as Wikipedia and LIGHT. \autoref{table:data_stats} presents dataset statistics; the full detailed descriptions and more information on how labels were inferred or imputed in \autoref{appendix:data}. 

Some of the datasets contain gender annotations provided by existing work. For example, classifiers trained for style transfer algorithms have previously annotated the gender of Yelp reviewers \cite{subramanian2018multiple}. In other datasets, we infer the gender labels. For example, in datasets where users are first assigned a \textit{persona} to represent before chatting, often the gender of the persona is pre-determined. In some cases gender annotations are not provided. In these cases, we sometimes impute the label if we are able to do so with high confidence. More details regarding how this is done can be found in \autoref{appendix:data}.

\paragraph{Collected Evaluation Dataset.}

We use a variety of datasets to train classifiers so they can be reliable on all dimensions across multiple domains. However, this weakly supervised data provides somewhat noisy training signal -- particularly for the \emph{masculine} and \emph{feminine} classes -- as the labels are automatically annotated or inferred. To enable reliable evaluation, we collect a specialized corpus, \newdataset, which acts as a gold-labeled dataset.

First, we collect conversations between two speakers. Each speaker is provided with a persona description containing gender information, then tasked with adopting that persona and having a conversation.\footnote{We note that  crowdworkers might perform genders in a non-authentic or idiosyncratic way when the persona gender doesn't match their gender. This would be an interesting avenue to explore in follow up work.} They are also provided with small sections of a biography from Wikipedia as the conversation topic. We observe that using biographies to frame the conversation encourages crowdworkers to discuss \emph{about/to/as} gender information.

To maximize the \emph{about/to/as} gender information contained in each utterance, we perform a second annotation over each utterance in the dataset. In this next phase, we ask annotators to rewrite each utterance to make it very clear that they are speaking \textsc{about} a man or a woman, speaking \textsc{as} a man or a woman, and speaking \textsc{to} a man or a woman. 
For example, given the utterance \textit{Hey, how are you today? I just got off work}, a valid rewrite to make the utterance \textsc{about} a woman could be: \textit{Hey, I went for a coffee with my friend and her dog} as the \textit{her} indicates a woman. A rewrite such as \textit{I went for a coffee with my friend} is not acceptable as it does not mention that the friend is a woman. 
After each rewritten utterance, evaluators label how confident they are that someone else would predict that the text is \textit{spoken about, spoken as, or spoken to} a man or woman. For the rewritten utterance \textit{I just got back from football practice}, many people would guess that the utterance was said by a man, as more men play football then women, but one cannot be certain (as women also play or coach football). An example instance of the task is shown in \autoref{table:new_data_example} and the interface is shown in Appendix Figure~\ref{fig:annotationfig}.

\begin{table}
\centering
\scalebox{0.95}{
\setlength\tabcolsep{3pt}
\begin{tabular}{lcccc}
\toprule
Model & \multicolumn{4}{c}{\textbf{Performance}} \\ 
      &  M & F & N & Avg. \\
\midrule 
Multi-Task & 87.4 & 86.65 & 55.2 & 77.22 \\ 
\midrule 
Wikipedia Only & 88.65 & 88.22 & 68.58 & 81.82 \\
 -gend words & 86.94 & 74.62 & 74.33 & 78.63 \\
 -gend words and names & 82.10 & 82.52 & 55.21 & 73.28 \\
\bottomrule
\end{tabular}
}
\caption{\label{table:wiki_variations} \textbf{Ablation of gender classifiers} on the Wikipedia test set. We report the model accuracy on the masculine, feminine, and neutral classes, as well as the average accuracy across them. We train classifiers (1) on the entire text (2) after removing explicitly gendered words using a word list and (3) after removing gendered words and names. While masking out gendered words and names makes classification more challenging, the model still obtains high accuracy.}
\end{table}

\section{Results}\label{sec:results}

\subsection{about/to/as Gender Classification}

\paragraph{Quality of Classification Models.} 

We compare models that classify along a single dimension compared to one that multitasks across all three. To enable high quality evaluation along our proposed three dimensions, we use \newdataset to evaluate. We measure the percentage accuracy for masculine, feminine, and neutral classes. We do not evaluate on the unknown class, as it is not modeled.
Classifier results on \newdataset~are shown in Table~\ref{table-one}. 

We find that the multitask classifier has the best average performance across all dimensions, with a small hit to single-task performance in the \emph{about} and \emph{as} dimensions.  As expected, the single task models are unable to transfer to other dimensions: this is another indication that gender information manifests differently along each dimension.
Training for a single task allows models to specialize to detect and understand the nuances of text that indicates bias along one of the dimensions. However, in a multitask setting, models see additional data along the other dimensions and can possibly learn to generalize to understand what language characterizes bias across multiple dimensions.

\paragraph{Performance by Dataset.} 
The gender classifiers along the \textsc{to, as} and \textsc{about} dimensions are trained on a variety of different existing datasets across multiple domains. We analyze which datasets are the most difficult to classify correctly in \autoref{table-two}. We find that \textsc{about} is the easiest dimension, particularly data from Wikipedia or based on Wikipedia, such as Funpedia and Wizard of Wikipedia, achieving almost 80\% accuracy. 

The \textsc{to} and \textsc{as} directions are both more difficult, likely as they involve more context clues rather than relying on textual attributes and surface forms such as \textit{she} and \textit{he} to predict correctly. We find that generally the datasets have similar performance, except Yelp restaurant reviews, which has a 70\% accuracy on predicting \textsc{as}. 

\paragraph{Analysis of Classifier Performance.} 
We break down choices made during classifier training by comparing different models on the Wikipedia (\textsc{about} dimension). We train a single classifier of \textsc{about}, and train with the variations of masking out gendered words and names. As gendered words such as \textit{her} and names are very correlated with gender, masking can force models into a more challenging but nuanced setting where they must learn to detect bias from the remaining text. We present the results in Table \ref{table:wiki_variations}. As expected, masking out gendered words and names makes it harder  to classify the text, but the model is still able to obtain high accuracy.

\begin{table}[t]
\begin{center}
\small
\begin{tabular}{p{9em}p{4em}p{4em}}
\toprule
\multicolumn{2}{l}{\textbf{Generation Statistics}} \\
\midrule
{\it Control Token} & {\it \# Gend. words} & {\it Pct. masc. }\\
\midrule 
{\sc TO}:feminine & 246 & 48.0 \\
{\sc AS}:feminine & 227 & 51.0 \\
{\sc ABOUT}:feminine & 1151 & 19.72 \\
Word list, feminine & 1158 & 18.22 \\
\midrule
{\sc TO}:masculine & 372 & 75.0 \\
{\sc AS}:masculine & 402 & 71.6 \\ 
{\sc ABOUT}:masculine & 800 & 91.62\\
Word list, masculine & 1459 & 94.8 \\
\bottomrule
\end{tabular}
\caption{\textbf{Word statistics} measured on text generated from 1000 different seed utterances from ConvAI2 for each control token, as well as for our baseline model trained using word lists. We measure the number of gendered words (from a word list) that appear in the generated text as well as the percentage of masculine-gendered words among all gendered words. Sequences are generated with top-$k$ sampling, $k=10$, with a beam size of $10$ and $3$-gram blocking.}
\label{table:generation_stats}
\end{center}
\end{table}

\section{Applications}\label{sec:applications}
We demonstrate the broad utility of our multi-task classifier by applying them to three different downstream applications. First, we show that we can use the classifier to control the genderedness of generated text. Next, we demonstrate its utility in biased text detection by applying it Wikipedia to find the most gendered biographies. Finally, we evaluate our classifier on an offensive text detection dataset to explore the interplay between offensive content and genderedness.

\subsection{Controllable Generation}\label{subsec:control_gen} 

By learning to associate control variables with textual properties, generative models can be controlled at inference time to adjust the generated text based on the desired properties of the user. This has been applied to a variety of different cases, including generating text of different lengths \cite{fan2017controllable}, generating questions in chit-chat \cite{see2019makes}, and reducing bias \cite{dinan2019queens}. 

Previous work in gender bias used word lists to control bias, but found that word lists were limited in coverage and applicability to a variety of domains \cite{dinan2019queens}. However, by decomposing bias along the \textsc{to, as, and about} dimensions, fine-grained control models can be trained to control these different dimensions separately. This is important in various applications --- for example, one may want to train a chatbot with a specific personality, leaving the \textsc{as} dimension untouched, but want the bot to speak to and about everyone in a similar way. In this application, we train three different generative models, each of which controls generation for gender along one of the \textsc{to}, \textsc{as}, and \textsc{about} dimensions. 

\paragraph{Methods} We generate training data by taking the multi-task classifier and using it to classify 250,000 textual utterances from Reddit, using a previously existing dataset extracted and obtained by a third party and made available on pushshift.io. This dataset was chosen as it is conversational in nature, but not one of the datasets that the classifier was trained on.  We then use the labels from the classifier to prepend the utterances with tokens that indicate gender label along the dimension. For example for the \textsc{about} dimension, we prepend utterances with tokens \emph{ABOUT:\textless gender\_label\textgreater}, where \emph{\textless gender\_label\textgreater} denotes the label assigned to the utterance via the classifier. At inference time, we choose control tokens to manipulate the text generated by the model. 

We also compare to a baseline for which the control tokens are determined by a word list: if an utterance contains more masculine-gendered words than feminine-gendered words from the word list it is labeled as \emph{masculine} (and vice versa for \emph{feminine}); if it contains no gendered words or an equal number of masculine and feminine gendered words, it is labeled as \emph{neutral}. Following \citet{dinan2019queens}, we use several existing word lists \cite{zhao-etal-2018-learning,zhao-etal-2019-gender,hoyle2019}.

For training, we fine-tune a large, Transformer sequence-to-sequence model pretrained on Reddit. At inference time, we generate text via top-$k$ sampling \cite{fan2018hierarchical}, with $k=10$ with a beam size of $10$, and $3$-gram blocking. We force the model to generate a minimum of $20$ BPE tokens. 

\paragraph{Qualitative Results.} Example generations from various control tokens (as well as the word list baseline) are shown in Table~\ref{table:generatedexamples} in the Appendix. These examples illustrate how controlling for gender over different dimensions yields extremely varied responses, and why limiting control to word lists may not be enough to capture these different aspects of gender. For example, adjusting \textsc{as} to `feminine' causes the model to write text such as \textit{Awwww, that sounds wonderful}, whereas setting \textsc{as} to masculine generates \textit{You can do it bro!} 

\paragraph{Quantitative Results.} Quantitatively, we evaluate by generating 1000 utterances seeded from ConvAI2 using both \emph{masculine} and \emph{feminine} control tokens and counting the number of gendered words from a gendered word list that also appear in the generated text.  Results are shown in Table~\ref{table:generation_stats}. 

Utterances generated using \emph{about} control tokens contain many more gendered words. One might expect this, as when one speaks \emph{about} another person, one may refer to them using gendered pronouns. We observe that for the control tokens \emph{TO:feminine} and \emph{AS:feminine}, the utterances contain a roughly equal number of masculine-gendered and feminine-gendered words. This is likely due to the distribution of such gendered words in the training data for the classifier in the \emph{to} and \emph{as} dimensions.
The ConvAI2 and Opensubtitles data show similar trends: on the ConvAI2 data, fewer than half of the gendered words in \emph{SELF:feminine} utterances are feminine-gendered, and on the Opensubtitles data, the ratio drops to one-third.\footnote{The Opensubtitles data recalls the Bechdel test, which asks ``whether a work [of fiction] features at least two women who talk to each other about something other than a man." \cite{wiki:bechdel}}
By design, the word list baseline has the best control over whether the generations contain words from this word list.  These results, as well as the previously described qualitative results, demonstrate why evaluating and controlling with word lists is insufficient --- word lists do not capture all aspects of gender.


\begin{table}[t]
\begin{center}
\small 
\begin{tabular}{lllll}
\toprule
\multicolumn{4}{l}{\textbf{Percentage of masculine-gendered text}} \\
\midrule 
\emph{Dim} & \emph{Safe} & \emph{Offensive} & \emph{t-statistic} & \emph{p-value} \\ 
{\sc ABOUT} & 81.03 & 70.66 &5.49  & 5.19e-08  \\
{\sc TO} & 44.68 & 60.15 & -22.02 &1.94e-46\\ 
{\sc AS} & 42.29 & 65.12 & -14.56 & 1.05e-99\\ 
\midrule 
\end{tabular}
\end{center}
\caption{\textbf{Genderedness of offensive content.} We measure the percentage of utterances in both the "safe" and "offensive" classes that are classified as \emph{masculine-gendered}, among utterances that are classified as either \emph{masculine-} or \emph{feminine-gendered}. We test the hypothesis that safe and offensive classes distributions of \emph{masculine-gendered} utterances differ using a $t$-test and report the p-value for each dimension. }
\label{table:off_content}
\end{table}

\subsection{Bias Detection}\label{subsec:bias} 

Creating classifiers along different dimensions can be used to detect gender bias in any form of text, beyond dialogue itself. We investigate using the trained classifiers to detect the most gendered sentences and paragraphs in various documents, and analyze what portions of the text drive the classification decision. Such methods could be very useful in practical applications such as detecting, removing, and rewriting biased writing. 

\paragraph{Methods.} We apply our classification models by detecting the most gendered biographies in Wikipedia.  We use the multitask model to score each paragraph among a set of $65,000$ Wikipedia biographies, where the score represents the probability that the paragraph is \emph{masculine} in the \emph{about} dimension. We calculate a \textit{masculine genderedness score} for the page by taking the median among all paragraphs in the page. 

\begin{table}[]
    \centering
    \begin{tabular}{lll}
    \toprule 
    \multicolumn{3}{l}{\textbf{Masculine genderedness scores}} \\ 
    \midrule 
    Biographies & Average & Median \\ 
    \midrule
    All & 0.74 & 0.98 \\
    Men & 0.90 & 0.99 \\ 
    Women & 0.042 & 0.00085 \\ 
    \bottomrule
    \end{tabular}
    \caption{\textbf{Masculine genderedness scores of Wikipedia bios.}~We calculate a \textit{masculine genderedness score} for a Wikipedia page by taking the median $p_x = P(x \in \mbox{ABOUT:masculine})$ among all paragraphs $x$ in the page, where $P$ is the probability distribution given by the classifier. We report the average and median scores for all biographies, as well as for biographies of men and women respectively.}
    \label{table:wiki_quant}
\end{table}

\paragraph{Quantitative Results.} We report the average and median \emph{masculine genderedness scores} for all biographies in the set of $65,000$ that fit this criteria, and for biographies of men and women in \autoref{table:wiki_quant}. We observe that while on average, the biographies skew largely toward \emph{masculine} (the average score is $0.74$), the classifier is more confident in the \emph{femininity} of pages about women than it is in the \emph{masculinity} of pages about men: the average \emph{feminine genderedness score} for pages about women is $1 - 0.042 = 0.958$, while the average \emph{masculine genderedness score} for pages about men is $0.90$.  This might suggest that biographies about women contain more gendered text on average.

\paragraph{Qualitative Results.} We show the pages---containing a minimum of 25 paragraphs---with the minimum score (most feminine-gendered biographies) and the maximum score (most masculine-gendered biographies) in \autoref{table:wiki_most_gendered} in the Appendix. We observe that the most masculine-gendered biographies are mostly composers and conductors, likely due to the historical gender imbalance in these occupations. Amongst the most feminine gendered biographies, there are many popular actresses from the mid-20th century. By examining the \emph{most} gendered paragraph in these biographies, anecdotally we find these are often the paragraphs describing the subject's life after retirement. For example, the most gendered paragraph in Linda Darnell's biography contains the line \textit{Because of her then-husband, Philip Liebmann, Darnell put her career on a hiatus,} which clearly reflects negative societal stereotypes about the importance of women's careers \citep{hiller1982,duxbury1991,pavalko1993,byrne2017,reid2018}.

\subsection{Offensive Content} 

Finally, the interplay and correlation between gendered text and offensive text is an interesting area for study, as many examples of gendered text---be they explicitly or contextually gendered---are disparaging or have negative connotations (e.g.,  ``cat fight'' and ``doll''). 
There is a growing body of research on detecting offensive language in text. In particular, there has been recent work aimed at improving the detection offensive language in the context of dialogue \cite{dinan2019build}.
We investigate this relationship by examining the distribution of labels output by our gender classifier on data that is labeled for offensiveness. 


\paragraph{Methods.} For this application, we use the \textit{Standard} training and evaluation dataset created and described in \citet{dinan2019build}. We examine the relationship between genderedness and offensive utterances by labeling the gender of utterances (along the three dimensions) in both the ``safe" and ``offensive" classes in this dataset using our multitask classifier.  We then measure the ratio of utterances labeled as \emph{masculine-gendered} among utterances labeled as either \emph{masculine-} or \emph{feminine-gendered}. 

\paragraph{Quantitative Results.} Results are shown in \autoref{table:off_content}.  We observe that, on the \emph{self} and \emph{partner} dimensions, the safe data is more likely to be labeled as \emph{feminine} and the offensive data is more likely to be labeled as \emph{masculine}. We test the hypothesis that these distributions are unequal using a T-test, and find that these results are significant.

\paragraph{Qualitative Results.} To explore how offensive content differs when it is \textsc{about} women and \textsc{about} men, we identified utterances for which the model had high confidence (probability $>0.70$) that the utterance was \emph{feminine} or \emph{masculine} along the \textsc{about} dimension. After excluding stop words and words shorter than three characters, we hand-annotated the top 20 most frequent words as being \textit{explicitly gendered}, a \textit{swear word}, and/or bearing \textit{sexual connotation}. For words classified as masculine, 25\% of the masculine words fell into these categories 
, whereas for words classified as feminine, 75\% of the words fell into these categories. 

\section{Conclusion}
We propose a general framework for analyzing gender bias in text by decomposing it along three dimensions: (1) gender of the person or people being spoken about (\textsc{about}), (2) gender of the addressee (\textsc{to}), and (2) gender of the speaker (\textsc{as}). We show that classifiers can detect bias along each of these dimensions. We annotate eight large existing datasets along our dimensions, and also contribute a high quality evaluation dataset for this task. We demonstrate the broad utility of our classifiers by showing strong performance on controlling bias in generated dialogue, detecting genderedness in text such as Wikipedia, and highlighting gender differences in offensive text classification.

\bibliography{anthology,emnlp2020}
\bibliographystyle{acl_natbib}

\appendix

\newpage \clearpage

\section{Existing Data Annotation}\label{appendix:data}

We describe in more detail how each of the eight training datasets is annotated:

\begin{enumerate}
    \setlength\itemsep{.1em}
    \item \textbf{Wikipedia} - to annotate \textsc{about}, we use a Wikipedia dump and extract biography pages. We identify biographies using named entity recognition applied to the title of the page \cite{spacy2}. We label pages with a gender based on the number of gendered pronouns (\textit{he} vs. \textit{she} vs. \textit {they}) and label each paragraph in the page with this label for the {\sc about} dimension.\footnote{This method of imputing gender is similar to the one used in \newcite[1142]{reagle2011} and \newcite{bamman2014}, except we also incorporate non-oppositional gender categories, and rely on basic counts without scaling.}
    Wikipedia is well known to have gender bias in equity of biographical coverage and lexical bias in noun references to women \citep{reagle2011, graells2015, wagner2015, klein2015, klein2016, wagner2016}, making it an interesting test bed for our investigation.
    \item \textbf{Funpedia} - Funpedia \cite{miller2017parlai} contains rephrased Wikipedia sentences in a more conversational way. We retain only biography related sentences and annotate similar to Wikipedia, to give {\sc about} labels. 
    \item \textbf{Wizard of Wikipedia} - Wizard of Wikipedia \cite{dinan2019wizard} contains two people discussing a topic in Wikipedia. We retain only the conversations on Wikipedia biographies and annotate to create {\sc about} labels.
    \item \textbf{ImageChat} - ImageChat \cite{shuster2018imagechat} contains conversations discussing the content of an image. We use the \cite{xu2015show} image captioning system\footnote{\url{https://github.com/AaronCCWong/Show-Attend-and-Tell}} to identify the contents of an image and select gendered examples. 
    \item \textbf{Yelp} - we use the Yelp reviewer gender predictor developed by \cite{subramanian2018multiple} and retain reviews for which the classifier is very confident -- this creates labels for the author of the review ({\sc as}). We impute {\sc about} labels on this dataset using a classifier trained on the datasets 1-4.
    \item \textbf{ConvAI2} - ConvAI2 \cite{dinan2019second} contains persona-based conversations. Many personas contain sentences such as \textit{I am a old woman} or \textit{My name is Bob} which allows annotators to annotate the gender of the speaker ({\sc as}) and addressee ({\sc to}) with some confidence. Many of the personas have unknown gender. We impute {\sc about} labels on this dataset using a classifier trained on the datasets 1-4.
    \item \textbf{OpenSubtitiles} - OpenSubtitles\footnote{http://www.opensubtitles.org/} \cite{lison2016opensubtitles2016} contains subtitles for movies in different languages. We retain English subtitles that contain a character name or identity. We annotate the character's gender using gender kinship terms such as \textit{daughter} and gender probability distribution calculated by counting the masculine and feminine names of baby names in the United States\footnote{\href{https://catalog.data.gov/dataset/baby-names-from-social-security-card-applications-national-level-data}{https://catalog.data.gov/dataset/baby-names-from-social-security-card-applications-national-level-data}}. Using the character's gender, we get labels for the {\sc as} dimension. We get labels for the {\sc to} dimension by taking the gender of the next character to speak if there is another utterance in the conversation; otherwise, we take the gender of the \emph{last} character to speak. We impute {\sc about} labels on this dataset using a classifier trained on the datasets 1-4.
    \item \textbf{LIGHT} - LIGHT contains persona-based conversation. Similarly to ConvAI2, annotators labeled the gender of each persona \cite{dinan2019queens}, giving us labels for the speaker ({\sc as}) and speaking partner ({\sc to}). We impute {\sc about} labels on this dataset using a classifier trained on the datasets 1-4.
\end{enumerate}

\section{New Evaluation Dataset}

\begin{figure*}[t]
    \centering
    \includegraphics[width=6in]{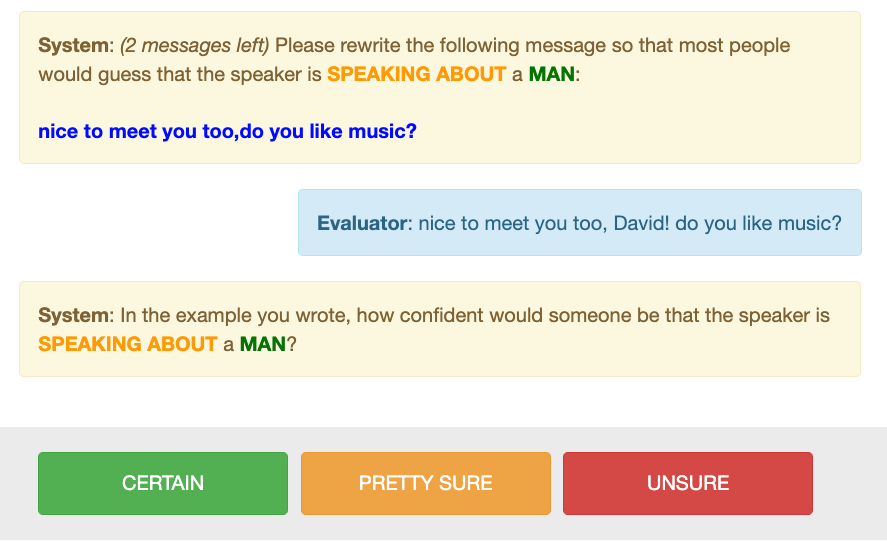}
    \caption{\textbf{Annotation interface}. Annotation interface for collecting \newdataset. Annotators were shown an utterance from a conversation, and asked to re-write it such that it is clear they would be speaker about/to/as a man or a woman. They were then asked for their confidence level.}
    \label{fig:annotationfig}
\end{figure*}

The interface for our new evaluation dataset \newdataset~ can be seen in \autoref{fig:annotationfig}. Examples from the new dataset can be found in \autoref{table:new_data_example}.

\begin{table*}
\setlength\tabcolsep{10pt}
\centering
\small 
\begin{tabular}{p{22em}p{3em}p{3em}p{5em}}
    \toprule
    Utterance & Dim. & Gender & Confidence \\ 
    \midrule
    \textbf{\emph{Original}}: That's interesting. I am a Chef. What are your hobbies \\
    \textbf{\emph{Rewrite}}:  that's interesting. i am a chef and nfl player what are your hobbies & {\sc as} & masc. & certain \\
    \midrule
    \textbf{\emph{Original}}: My name is Zachery but most call me Z. What's your name miss? \\
    \textbf{\emph{Rewrite}}: My name is Tina but most call me T. What's your name miss? & {\sc as} & fem. & pretty sure \\
    \midrule
    \textbf{\emph{Original}}: I said I like to wrestle with my kids for fun. What do you do for work?  \\
    \textbf{\emph{Rewrite}}: i said i like to wrestle with my son for fun. what do you do for work? & {\sc about} & masc. & certain \\
    \midrule
    \textbf{\emph{Original}}: just checking up on you by saying hi  \\
    \textbf{\emph{Rewrite}}: Hanna said she was just checking up on you by saying hi & {\sc about} & fem. & pretty sure  \\
    \midrule 
    \textbf{\emph{Original}}: How are you doing today Jahsim? \\
    \textbf{\emph{Rewrite}}: How are you doing today Mr. Jahsim? & {\sc to} & masc. & certain \\
    \midrule
    \textbf{\emph{Original}}: My name is Alexsandro and I am a bee farmer! \\
    \textbf{\emph{Rewrite}}:  My name is alexsandro and I am a bee farmer! How about you ma'am? & {\sc as} & fem. & certain \\
    \bottomrule
\end{tabular}
\caption{\textbf{Examples from the \newdataset}. Crowdworkers were asked to re-write dialogue utterances such that most people would guess that the utterance was either said to, said by, or about a man or a woman. Afterwards, they were asked to give a confidence level in their re-write, meant to capture the differences between statistical biases (more men play football than women) and fact (you do not have to be a man to play football).}
\label{table:new_data_example}
\end{table*}

\section{Applications}

Example generations for various control tokens, as well as for our word list baseline, are shown in \autoref{table:generatedexamples}. See \autoref{subsec:control_gen} on Controllable Generation in the main paper for more details.

The top 10 most gendered Wikipedia biographies are shown in \autoref{table:wiki_most_gendered}. See \autoref{subsec:bias} on Detecting Bias in the main paper for more details.

\begin{table*}[t]
\setlength\tabcolsep{3pt}
\begin{center}
\small
\begin{tabular}{p{50em}}
\toprule
\textbf{Input:} ~ Awww! Cute! Our are a handful, but they are my babies! 4 wiener dogs. \\
\textbf{TO:feminine} ~ I can't wait for your children to get married.. I want one so badly! : ) \\
\textbf{AS:feminine} ~ Awwww, that sounds wonderful. Can't wait to see them in person! : ) \\
\textbf{ABOUT:feminine} ~ You're my hero. That's a cute dog. How long has she been with you? \\
\textbf{Word list, feminine} ~ That's pretty awesome! She was a little bit overweight but she's a sweet girl now \\
\midrule 
\textbf{Input:} ~ Does vegan mean you do not want to be a super hero? \\
\textbf{TO:feminine} ~ I'd love to be vegan, but it just isn't something I have any interest in doing. \\
\textbf{AS:feminine} ~ I don't want to become super heroes! I just think it would be cool to be vegan. \\
\textbf{ABOUT:feminine} ~ She's a girl. She can't be a superhero!!! 11! 1! 1 1! \\
\textbf{Word list, feminine} ~ It does. She is a vegan and I am a super - hero. I am not vegan, and I don't want to become any sort of superhero. \\
\midrule 
\textbf{Input:} ~ That is an good. Maybe you can make friends in a club or on a job? \\
\textbf{TO:feminine} ~ I was going to do that, but then I realized I didn't want to be alone. \\
\textbf{AS:feminine} ~ No, I've been working. I don't think I can make friendships online. \\
\textbf{ABOUT:feminine} ~ I think the problem is she's a girl, so there's not a lot of opportunity to make friends. \\
\textbf{Word list, feminine} ~ I have a girlfriend and I live in a small town. I don't know what clubs she can make any friends in. \\
\midrule 
\midrule
\textbf{Input:} ~ Its been 2 years I am a grease monkey and been trying to stay busy \\
\textbf{TO:masculine} ~ I'm not sure this is true. I've seen a lot of grease monkeys on reddit. I don't think there's any evidence it's true. \\
\textbf{AS:masculine} ~ You can do it bro! Just keep at it! You got this!!! : d ) \\
\textbf{ABOUT:male} ~ I'd love to be a grease guy but I just can't make it to the gym. : ( \\
\textbf{Word list, masculine} ~ You're not alone bro. I'm trying to be that guy, but it's a little hard sometimes. \\
\midrule
\textbf{Input:} ~ Have you ever almost died? I nearly did from urinary cancer when I was 18. \\
\textbf{TO:masculine} ~ No one in my family has died. I have only seen a handful of people die. I am sure you will be fine. \\
\textbf{AS:masculine} ~ No I haven't. But it's a lot better now, thanks to all the help I've received. \\
\textbf{ABOUT:masculine} ~ He said * nearly *. It's more likely to happen when you're older than the average person. \\
\textbf{Word list, masculine} ~ I've only been alive for 3 days. I have no idea what I'm going to do next, but I'll be okay. \\
\midrule
\textbf{Input:} ~ I totally believe you. I love working out. \\
\textbf{TO:masculine} ~ I'm going to start doing this, thanks! If it works, I'll post it on r / fitness! \\
\textbf{AS:masculine} ~ Thanks! I've always been really into exercise, too, and it's definitely helped me. \\
\textbf{ABOUT:masculine} ~ Thank you! I'm really glad he does : ). I've been working out since october. \\
\textbf{Word list, masculine} ~ Me too! I love being able to go to the gym without feeling like I'm just a little kid. It's so rewarding when you get back in the swing of things. \\
\bottomrule
\end{tabular}
\end{center}
\caption{\textbf{Example generations} from a generative model trained using controllable generation, with control tokens determined by the classifier. Sequences are generated with top-$k$ sampling, $k=10$, with a beam size of $10$ and $3$-gram blocking. Input is randomly sampled from the ConvAI2 dataset.}
\label{table:generatedexamples}
\end{table*}

\begin{table*}[t]
\setlength\tabcolsep{10pt}
\begin{center}
\small
\begin{tabular}{p{22em}p{22em}}
\toprule
\emph{Most Feminine} & \emph{Most Masculine}\\
\midrule
1. \textbf{Edie Sedgwick:} was an American actress and fashion model... & 1. \textbf{Derek Jacobi:} is an English actor and stage director... \\
2. \textbf{Linda Darnell:} was an American film actress... & 2. \textbf{Bohuslav Martinů:} was a Czech composer of modern classical music... \\
3. \textbf{Maureen O'Hara:} was an Irish actress and singer... & 3. \textbf{Carlo Maria Giulini:} was an Italian conductor... \\
4. \textbf{Jessica Savitch:} was an American television news presenter and correspondent,... & 4. \textbf{Zubin Mehta:} is an Indian conductor of Western classical music... \\
5. \textbf{Patsy Mink:} Mink served in the U.S. House of Representatives... & 5. \textbf{John Barbirolli:} was a British conductor and cellist ... \\
6. \textbf{Shirley Chisholm:} was an American politician, educator, and author... & 6. \textbf{Claudio Abbado:} was an Italian conductor... \\
7. \textbf{Mamie Van Doren:} is an American actress, model, singer, and sex symbol who is... & 7. \textbf{Ed Harris:} is an American actor, producer, director, and screenwriter... \\
8. \textbf{Jacqueline Cochran:} was a pioneer in the field of American aviation and one of t... & 8. \textbf{Richard Briers:} was an English actor... \\
9. \textbf{Chloë Sevigny:} is an American actress, fashion designer, director, and form... & 9. \textbf{Artur Schnabel:} was an Austrian classical pianist, who also composed and tau... \\
10. \textbf{Hilda Solis:} is an American politician and a member of the Los Angeles Co... & 10. \textbf{Charles Mackerras:} was an Australian conductor... \\
\bottomrule
\end{tabular}
\end{center}
\caption{\textbf{Most gendered Wikipedia biographies} We ran our multi-task classifier over 68 thousand biographies of Wikipedia. After selecting for biographies with a minimum number of paragraphs (resulting in 15.5 thousand biographies) we scored them to determine the most \emph{masculine} and \emph{feminine} gendered.}.
\label{table:wiki_most_gendered}
\end{table*}

\end{document}